\documentclass[lettersize,journal]{IEEEtran}

\usepackage[letterpaper, top=3.2cm, bottom=3.2cm, left=1.92cm, right=1.92cm]{geometry}
\usepackage{multirow}
\usepackage[table,xcdraw]{xcolor}
\usepackage{graphicx}
\usepackage{amsmath}
\usepackage{amsfonts}
\usepackage{algorithm}
\usepackage{amsthm,amssymb}
\usepackage{mathrsfs}
\usepackage{algorithmic}
\usepackage{makecell}
\usepackage{easyReview}
\usepackage{booktabs}
\setcellgapes{3pt}
\usepackage{verbatim}
\setcellgapes{3pt}

\usepackage[justification=centering]{caption}
\usepackage{color}
\usepackage[subfigure]{tocloft}
\usepackage{subfigure}
\usepackage{tabularx}
\usepackage{amsmath, bm}
\usepackage{hyperref}
\hypersetup{hidelinks,
	colorlinks=true,
	allcolors=blue,
	pdfstartview=Fit,
	breaklinks=true}

\usepackage{cite}

\ifCLASSINFOpdf
\else
\fi
\begin{document}

%
\title{Semantic Communications with Computer Vision Sensing for Edge Video Transmission}

\author{Yubo Peng, \textit{Student Member, IEEE}, Luping Xiang, \textit{Member, IEEE}, Kun Yang, \textit{Fellow, IEEE}, Kezhi Wang, \textit{Senior Member, IEEE}, and Mérouane Debbah, \textit{Fellow, IEEE} 
	\thanks{
	
	Yubo Peng (ybpeng@smail.nju.edu.cn), Luping Xiang (luping.xiang@nju.edu.cn), and Kun Yang (kunyang@essex.ac.uk) are with the State Key Laboratory of Novel Software Technology, Nanjing University, Nanjing, China, and the School of Intelligent Software and Engineering, Nanjing University (Suzhou Campus), Suzhou, China.
	
	Kezhi Wang (Kezhi.Wang@brunel.ac.uk) is with the Department of Computer Science, Brunel University London, UK.

    Mérouane Debbah (merouane.debbah@ku.ac.ae) is with the Department of Electrical Engineering and Computer Science and the KU 6G Center, Khalifa University, Abu Dhabi 127788, UAE.
	}
}

\markboth{Submitted for Review}%
{Shell \MakeLowercase{\textit{et al.}}: Bare Demo of IEEEtran.cls for IEEE Journals}
%



\maketitle


\begin{abstract}
Despite the widespread adoption of vision sensors in edge applications, such as surveillance, the transmission of video data consumes substantial spectrum resources. Semantic communication (SC) offers a solution by extracting and compressing information at the semantic level, preserving the accuracy and relevance of transmitted data while significantly reducing the volume of transmitted information. However, traditional SC methods face inefficiencies due to the repeated transmission of static frames in edge videos, exacerbated by the absence of sensing capabilities, which results in spectrum inefficiency.
To address this challenge, we propose a SC with computer vision sensing (SCCVS) framework for edge video transmission. The framework first introduces a compression ratio (CR) adaptive SC (CRSC) model, capable of adjusting CR based on whether the frames are static or dynamic, effectively conserving spectrum resources.
Additionally, we implement an object detection and semantic segmentation models-enabled sensing (OSMS) scheme, which intelligently senses the changes in the scene and assesses the significance of each frame through in-context analysis. Hence, The OSMS scheme provides CR prompts to the CRSC model based on real-time sensing results.
Moreover, both CRSC and OSMS are designed as lightweight models, ensuring compatibility with resource-constrained sensors commonly used in practical edge applications. Experimental simulations validate the effectiveness of the proposed SCCVS framework, demonstrating its ability to enhance transmission efficiency without sacrificing critical semantic information.

\end{abstract}


\section{Introduction}
\subsection{Backgrounds}
With the rapid advancement of the Internet of Things, vision sensors are increasingly deployed to provide intelligent services, particularly in the surveillance domain. Video surveillance is highly valued not only for its ability to solve crimes but also for its potential role in crime prevention. As a result, numerous vision sensors are commonly installed in public spaces, malls, and residential areas \cite{8493895}. Compared to wired networks, wireless-based sensors offer greater flexibility in deployment, especially in geographically dispersed or complex environments, as they eliminate the need for physical cables. However, while wireless transmission is more versatile, transmitting edge video, which typically involves large data sizes, imposes substantial spectrum resource demands. This challenge hinders the development of wireless sensor-based edge applications.

Semantic communication (SC), a key technology for 6G, significantly reduces the data transmission requirements by extracting and compressing information at the semantic level, while maintaining precision and relevance \cite{10558819,10670195}. Unlike traditional communication approaches, which prioritize error-free symbol delivery, SC focuses on achieving ``semantic fidelity," effectively mitigating the ``cliff effect" caused by decreasing signal-to-noise ratios (SNR) \cite{9679803}. Consequently, SC offers a promising solution to the issue of spectrum scarcity. However, many practical scenes remain static for long periods, resulting in numerous redundant frames within the video. Since traditional SC cannot sense scene changes, it applies a uniform compression ratio (CR) across all frames, leading to spectrum inefficiencies by unnecessarily transmitting redundant information.

With the growing demand for higher communication rates, the wireless industry is pushing carrier frequencies into bands traditionally reserved for radar sensing systems \cite{8828016}. Radar sensing technology can provide high-accuracy sensing for various applications \cite{9705498}, including autonomous vehicle driving, robot navigation, and indoor localization for virtual reality \cite{luo2023reconfigurable}. Therefore, radar sensing appears to offer a viable solution for detecting changes in scenes. However, this approach requires sensors to be equipped with advanced radar systems, which is difficult to implement with the widely deployed general-purpose vision sensors due to the high cost.

\subsection{Challenges}
Given the above background, several challenges arise in applying sensing-assisted SC for edge video transmission:
\begin{enumerate}
    \item \textit{Low communication efficiency}: 
    Most edge scenes are predominantly static, with dynamic changes occurring infrequently. This leads to the transmission of static frames, as SC cannot dynamically adapt the CR in real time based on scene content. Applying a uniform CR across the entire video results in inefficient use of spectrum resources.

     \item \textit{High sensing cost}:
    While radar sensing provides a promising solution for detecting changes in scenes, the high cost and complexity of radar systems, along with the need for sophisticated signal processing, make it difficult to implement on general-purpose vision sensors widely used for edge applications.

    \item \textit{Resource constraints of sensors}:
    Whether using artificial intelligence (AI)-driven SC or radar-based sensing, deploying such technologies on vision sensors presents challenges. General-purpose sensors, designed for widespread deployment, typically feature low-cost hardware with limited resources. Therefore, it is critical to balance performance requirements, such as low latency, with the constraints of limited sensor resources.
\end{enumerate}

\subsection{Contributions}
Recently, the rapid development of deep learning, particularly in the field of computer vision (CV), has led to significant advancements in perception technologies. CV models, such as Yolov10 \cite{wang2024yolov10} and FastSAM \cite{zhao2023fast}, can automatically extract features from large datasets and perform complex visual tasks such as image recognition, object detection, and semantic segmentation. Compared to traditional methods, CV offers substantial improvements in accuracy and robustness while running efficiently on standard hardware \cite{xie2021deep}. These advancements offer new opportunities for implementing cost-effective sensing systems, particularly in resource-constrained environments.
Therefore, we propose a SC with CV sensing (SCCVS) framework to enable efficient transmission of edge videos. Our main contributions are as follows:
\begin{enumerate}[]
	\item We develop a CR adaptive SC (CRSC) model for edge video transmission. Specifically, we design two distinct compression strategies of semantic representations: one to perform high-compression semantic encoding for static frames, as identified by CV sensing, and another to apply low-compression semantic encoding for the valuable scene frames. Additionally, during the training process, we introduce a knowledge distillation (KD)-based approach that enables the two SC models with different CRs of semantics to learn from each other, thereby enhancing the semantic quality of transmission, particularly under high-compression conditions. Thus, we solve the proposed first challenge.
	
	\item We propose an object detection and semantic segmentation models-enabled sensing (OSMS) scheme, leveraging object detection and semantic segmentation models to intelligently sense changes in edge scenes. OSMS provides CR prompts to the CRSC model based on its real-time analysis of each video frame. This CV-based sensing approach reduces the need for costly radar equipment, addressing the second challenge related to high sensing costs.
	
	\item We adopt lightweight AI models in the SCCVS framework to accommodate the resource constraints of vision sensors. Specifically, in the CRSC model, we use the Bilateral Transformer (BiFormer) \cite{zhu2023biformer} for semantic extraction and Kolmogorov-Arnold Networks (KAN) \cite{liu2024kan} for semantic encoding, both of which significantly reduce the parameter size of the semantic encoder. In OSMS, we employ Yolov10 for object detection and FastSAM for semantic segmentation, both capable of real-time inference on general-purpose sensors. This ensures the system operates within the resource limitations of the sensors, addressing the third challenge of limited resources.
    
	\item 
     We conduct experimental simulations on VIRAT Video Dataset \cite{oh2011large} and the results show that the proposed approach reduces data transmission by approximately 90\%, compared to the baseline method. This validates the effectiveness of the proposed framework.
\end{enumerate}

The structure of this paper is as follows. Section II introduces the related works. Section III provides a detailed description of the system model. 
Section IV presents the proposed SCCVS framework, which mainly includes the CRSC and OSMS schemes. 
Section V employs experimental simulations to evaluate the performance of the proposed methods. Lastly, Section VI concludes this paper.

\section{Related Works}
This section introduces the related works about SC for video transmission and CV for sensing, highlighting the differences between the existing works and ours. 

\subsection{SC for Video Transmission}
Given the unique advantages of SC in spectrum utilization, many researchers have explored SC for video transmission. Jiang \textit{et al.} \cite{9955991} proposed a semantic video conferencing network based on keypoint transmission, significantly reducing transmission resources with only minor losses in detailed expressions. Wang \textit{et al.} \cite{9953110} designed a deep video semantic transmission framework for end-to-end video transmission over wireless channels, utilizing nonlinear transforms and conditional coding to adaptively extract and transmit semantic representations across video frames using deep joint source-channel coding (JSCC). Tian \textit{et al.} \cite{10622487} introduced a synchronous SC system for video and speech transmission, using real-time facial transmission as a key use case. Bao \textit{et al.} \cite{10685066} developed a model division video SC (MDVSC) framework for efficient video transmission over noisy wireless channels, leveraging model division multiple access (MDMA) to extract common semantic representations from video frames and employing deep JSCC to handle channel noise. Eteke \textit{et al.} \cite{10647591} proposed a real-time semantic video communication method for general scenes, combining lossy semantic map coding with motion compensation to reduce bit rates while maintaining both perceptual and semantic quality.

\renewcommand{\arraystretch}{1.2} 
\begin{table}[htbp]
	\centering
	\caption{Comparison of our contributions with related literature on video SC}
	\label{tab:compare}
	\begin{tabular}{|c|c|c|c|c|c|c|}
\hline
Contributions                                                                  & Ours & \cite{9955991} & \cite{9953110} & \cite{10622487} & \cite{10685066} & \cite{10647591} \\ \hline
Video transmission                                                             & \checkmark    & \checkmark        & \checkmark        & \checkmark        & \checkmark        & \checkmark        \\ \hline
Video sensing                                                                  & \checkmark    & \checkmark        &          & \checkmark        &          &          \\ \hline
\begin{tabular}[c]{@{}c@{}}KD-based\\ model training\end{tabular}              & \checkmark    &          &          &          &          &          \\ \hline
\begin{tabular}[c]{@{}c@{}}Adaptive semantic\\ compression\end{tabular}        & \checkmark    &          & \checkmark        &          &          & \checkmark        \\ \hline
\begin{tabular}[c]{@{}c@{}}Bi-Former-based\\semantic extraction\end{tabular} & \checkmark    &          &          &          &          &          \\ \hline
\begin{tabular}[c]{@{}c@{}}KAN-based\\semantic encoding\end{tabular}          & \checkmark    &          &          &          &          &          \\ \hline
\end{tabular}
\end{table}

We compare the contributions of the above works with ours in Table \ref{tab:compare}. In summary, while these works present various approaches to improving video SC performance, they do not address the challenge of sensing changes in video content. As a result, these methods require the transmission of entire videos, leading to lower spectrum utilization. In contrast, our proposed SCCVS framework leverages sensing to identify changes in edge videos, enabling the SC model to optimize data transmission and achieve higher spectrum efficiency.

\subsection{CV for Sensing}
In the new generation of mobile communication systems, CV has emerged as a key enabler for proactive perception in wireless communication, such as locating mobile users and assisting with beam alignment \cite{lu2024semantic}, attracting wide interests of researchers. Bai \textit{et al}. \cite{9540364} proposed a novel visible light communication (VLC)-assisted perspective-four-line algorithm (V-P4L), which integrates VLC and CV techniques to achieve high localization accuracy, regardless of variations in LED height. Fusco \textit{et al}. \cite{10.1145/3371300.3383345} combined CV with existing informational signs, such as exit signs, along with inertial sensors and a 2D map to estimate and track user location within an environment. Wang \textit{et al}. \cite{8908665} introduced RF-MVO, a CV-assisted radio frequency identification (RFID) hybrid system for stationary RFID localization in 3D space, utilizing a lightweight 2D monocular camera mounted parallel to two reader antennas. Wen \textit{et al}. \cite{wen2023vision} developed a CV-aided millimeter-wave beam selection method that leverages environmental semantics extracted from user-side camera images to identify features affecting wireless signal propagation and accurately estimate the location of scatterers. Feng \textit{et al}. \cite{zhao2024environment} employed a pre-trained CV model to predict optimal beam configurations based on current environmental information.

In contrast to these existing approaches that typically analyze single frames, our work applies CV to sense the context of videos, thereby assisting SC in achieving higher transmission efficiency. Moreover, we also consider the lightweight of the CV models employed, ensuring they meet the resource constraints of the sensors utilized.

\section{System Model}
\begin{figure*}[htbp]
	\centering
	\includegraphics[width=15cm]{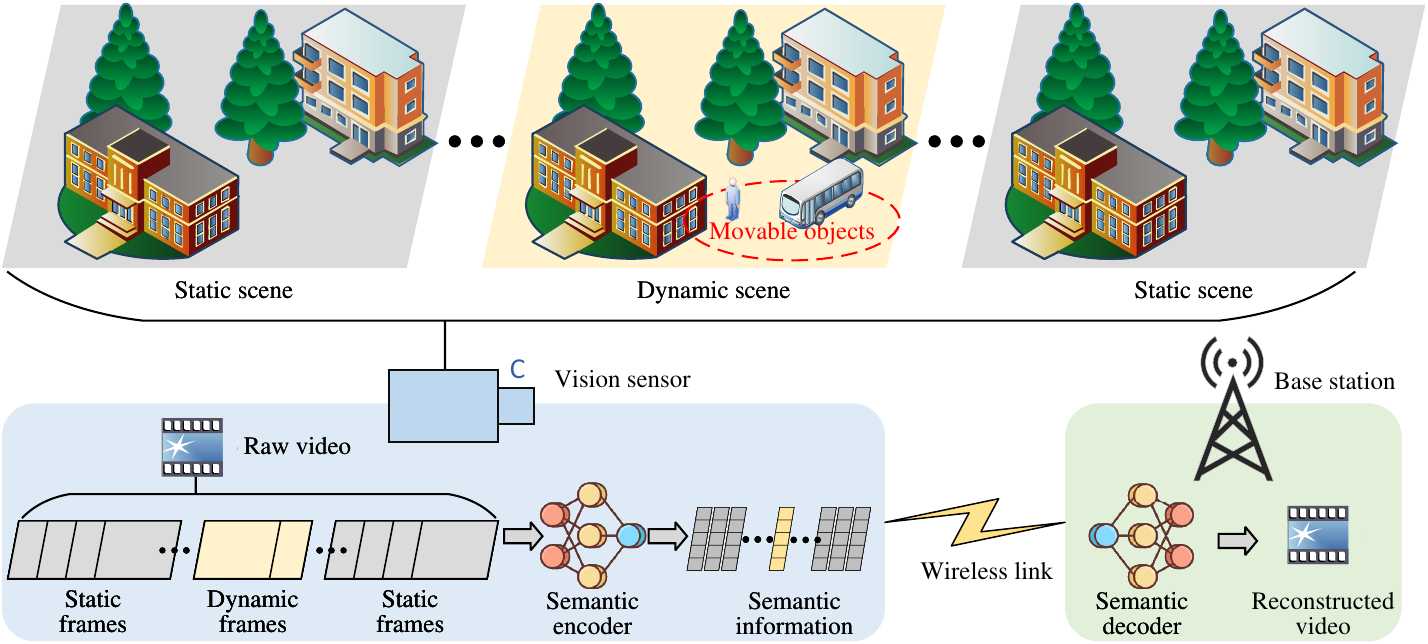}
	\caption{The system model of SC with sensing for transmitting edge videos.}
	\label{fig:system}
\end{figure*}
As illustrated in Fig. \ref{fig:system}, a video surveillance scenario is considered where the visual sensor has a fixed camera angle, such as in a parking lot, resulting in a static monitoring scene. The primary variations within the scene are due to movable objects like pedestrians and vehicles. These objects, however, are typically in motion for only a small fraction of the time and remain stationary for the majority of the observation period. Due to limited local storage capacity, the sensor must transmit the captured video data to a nearby base station (BS) via a wireless link. To address the issue of spectrum scarcity, a SC system with sensing capabilities at both the sensor and the BS is implemented for data transmission. In this system, the sensor functions as the transmitter, while the BS serves as the receiver. A semantic encoder is deployed at the sensor to extract and encode semantic information from the raw video data, and a semantic decoder is employed at the BS to decode the information and reconstruct the video. This approach transmits only high-density semantic information instead of raw video data, significantly reducing the bandwidth requirements.

\subsection{Semantic Communication Model}
In the proposed SC model, the captured raw video is transmitted frame-by-frame. This scenario is modeled as a point-to-point wireless image transmission system enabled by deep JSCC. The raw video is denoted as $\mathcal{V}=\{\mathbf{x}_i | i \in \{1,\dots, V\}\}$, where $\mathbf{x}_i$ represents the $i$-th frame, and $V$ is the total number of frames. The primary objective is to reconstruct all video frames at the receiver (i.e., the base station) transmitted by the transmitter (i.e., the vision sensor) under varying channel SNR and CR conditions.

\subsubsection{Encoder}
Each frame is assumed to have a height $H$, width $W$, and depth $C$. The source bandwidth of each frame is defined as $m = H \times W \times C$, leading to a total source bandwidth of $m \cdot V$ for the entire raw video.
At the transmitter, a semantic encoder coupled with signal modulation transforms the $i$-th frame $\mathbf{x}_i$ into an $n_i$-dimensional complex vector $\mathbf{c}_i \in \mathbb{C}^{n}_i$. This process is formulated as: \begin{equation}\label{eq:SC1}
 \mathbf{c}_i=F_\text{se}(\mathbf{x}_i,r_i, \alpha),
 \end{equation} 
 where $F_{\text{se}}(\cdot)$ denotes the semantic encoder with parameters $\alpha$, and $r_i = (m - n_i)/m$ is the CR for frame $\mathbf{x}_i$ \cite{kurka2020deepjscc}. Accordingly, the overall CR of the edge video is expressed as: \begin{equation}\label{eq:SC2}
    r = \frac{\sum_{i=1}^{V}r_i}{V}.
 \end{equation}

\subsubsection{Wireless channel}
When transmitted over a wireless fading channel, the complex vector $\mathbf{c}_i$ is subject to transmission impairments, including distortion and noise. This transmission process can be modeled as:
\begin{equation}\label{eq:SC3}
	\mathbf{y}_i = \mathbf{H}\cdot\mathbf{c}_i+\mathbf{N},
\end{equation}
where $\mathbf{y}_i$ is the received complex vector, $\mathbf{H}$ represents the channel gain between the transmitter and receiver, and $\mathbf{N}$ denotes the Additive White Gaussian Noise (AWGN). To enable end-to-end training of both the encoder and decoder, the channel model must support backpropagation. Consequently, the wireless channel is simulated using neural network-based approaches \cite{jiang2024large}. 

\subsubsection{Decoder}
Upon receiving the vector $\mathbf{y}_i$, the semantic decoder is responsible for reconstructing the corresponding frame. This reconstruction process can be expressed as: 
\begin{equation}\label{eq:SC4}
\hat{\mathbf{x}}_i = F_\text{sd}(\mathbf{y}_i,r_i,\beta),
\end{equation}
where $F_\text{sd}(\cdot)$ denotes the semantic decoder parameterized by $\beta$, and $\hat{\mathbf{x}}_i$ represents  the reconstructed $i$-th frame.
Upon completing the transmission of all frames, the reconstructed video, denoted as $\hat{\mathcal{V}}$, is obtained.
\subsection{Delay model}
During the uplink transmission of the complex vector from the vision sensor to the base station (BS), the transmission rate can be expressed as:
\begin{equation}
	v = B\log _{2}\left(1+\phi\right),
\end{equation}
where $B$ indicates the bandwidth and $\phi$ denotes the SNR. The transmission delay for the $i$-th frame is then given by:
\begin{equation}
	t_i=\frac{Z\left(\mathbf{c}_i\right)}{v},
\end{equation}
where $Z(\mathbf{c}_i)$ represents the number of bits required to transmit the complex vector $\mathbf{c}_i$ to the BS. Consequently, the total transmission delay for the edge video can be calculated as:
\begin{equation}
    T = \sum_{i=1}^{V} t_i.
\end{equation}

\subsection{Problem formulation}
Considering that edge video frames often contain high levels of redundancy with limited valuable information, traditional metrics that assess the consistency of every frame between the raw and reconstructed videos may not be appropriate. To more accurately assess the performance of video SC, it is essential to focus on minimizing differences in valuable frames while accounting for transmission delays. Thus, the objective function of the proposed SC system for edge video can be formulated as:
\begin{subequations}\label{eq:problem}
	\begin{align}
		\min_{\alpha, \beta,\mathcal{R}}\frac{1}{V}\sum_{i=1}^V (\mathbf{x}_i-\mathbf{\hat{x}}_i)^2 \cdot (1-r_i) + \zeta T,
	\end{align} 
	\begin{alignat}{1}
		\text{s.t.~} 
		& r_i \in \left\{ 0,1 \right\},~\forall i \in \{1,\dots,V\},
	\end{alignat}
\end{subequations}
where $\mathcal{R} = \{r_i|i=1,\dots,V\}$ represents the set of CR for each frame, and $\zeta$ denotes an adjustment coefficient. The constraint in Eq. (\ref{eq:problem}b) indicates that each frame is either compressed or not. 

To solve the optimization problem in Eq. (\ref{eq:problem}a), we propose the SCCVS framework. 
On the one hand, the CRSC module is designed to minimize the distortion term $\left(\mathbf{x}_{i}-\hat{\mathbf{x}}_{i}\right)^{2}$ by optimizing the parameters $\alpha$ and $\beta$ during model training. On the other hand, the OSMS module dynamically senses frame changes to optimize $r_i$, thereby reducing the transmission delay associated with static frames.


\section{Proposed SCCVS Framework}
\subsection{Overview}
\begin{figure*}[htbp]
	\centering
	\includegraphics[width=16cm]{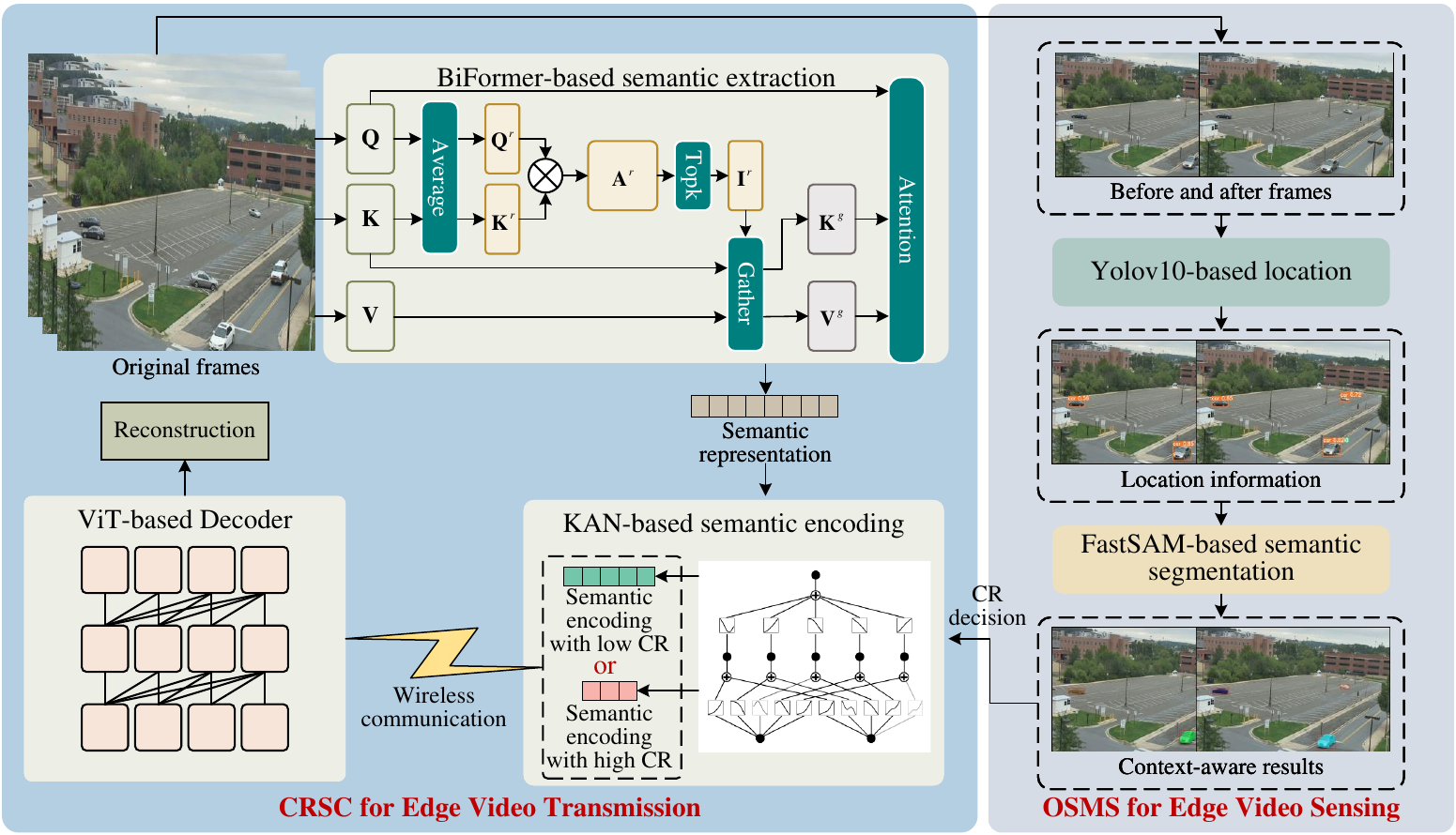}
	\caption{Illustration of the proposed SCCVS framework.}
	\label{fig:SCCVS}
\end{figure*}
In practical scenarios, addressing the substantial spectrum resource consumption caused by video transmission from vision sensors is crucial. To this end, we introduce the SCCVS framework, which integrates SC and CV sensing technologies to achieve efficient video transmission. As depicted in Fig. \ref{fig:SCCVS}, the framework consists of two primary modules:

\subsubsection{CRSC for Edge Video Transmission}
For each video frame, the CRSC module first utilizes a BiFormer to extract semantic information from a given frame $\mathbf{x}_i$, generating a high-dimensional semantic representation $\mathbf{s}_i$. A KAN is then employed to compress $\mathbf{s}_i$ based on the specified CR $r_i$, resulting in a semantic encoding $\mathbf{e}_i$ with either a high or low CR. This semantic encoding is subsequently modulated into a complex vector $\mathbf{c}_i$ for wireless transmission. At the receiver side, the complex vector $\mathbf{y}_i$, which may include noise distortions, is demodulated and processed by a KAN and ViT-based semantic decoder to reconstruct the frame $\hat{\mathbf{x}}_i$. The CRSC scheme is detailed in \textbf{Algorithm \ref{alg:CRSC1}}. Upon completing the transmission of all frames, the reconstructed video $\hat{\mathcal{V}}$ is obtained. Through this innovative SC approach, the CRSC module achieves highly efficient data transmission for edge video.

\subsubsection{OSMS for Edge Video Sensing}
To accurately detect changes within the scene and dynamically adjust the CR in the CRSC module, while minimizing the sensing cost for vision sensors, the OSMS module employs CV-based models to analyze video frames. Specifically, the framework employs Yolov10 to detect movable objects, such as pedestrians and vehicles, within the frames. FastSAM is then applied to segment key targets, isolating their corresponding pixel sets. The identified elements are subsequently analyzed across frames to detect contextual changes in the scene. Based on these results, the CR $r_i$ for the current frame is determined and fed into the CRSC module to guide the semantic encoding process. \textbf{Algorithm \ref{alg:OSMS}} provides a detailed description of the OSMS scheme. By leveraging CV-based image processing techniques, the OSMS module effectively reduces the high sensing costs typically associated with radar-based equipment.

To facilitate understanding, the workflow of the proposed SCCVS framework is described in \textbf{Algorithm \ref{alg:SCCVS}}.
\begin{algorithm}
	\caption{SCCVS Framework Workflow}
	\label{alg:SCCVS}
	\begin{algorithmic}[1]
		\REQUIRE $\mathcal{V}$.
		\ENSURE $\hat{\mathcal{V}}$.
		\FOR{$i = 1,2,\dots, V$}
        \STATE{Sense the current frame $\mathbf{x}_i$ and determine the corresponding CR $r_i$ using \textbf{Algorithm \ref{alg:OSMS}}.}
		\STATE{Reconstruct the frame $\hat{\mathbf{x}}_i$ based on \textbf{Algorithm \ref{alg:CRSC1}} and the determined CR $r_i$ .}
		\ENDFOR
        \STATE{Combine all reconstructed frames to obtain the final reconstructed video $\hat{\mathcal{V}}$.}
        \STATE{Assess the transmission quality of the video SC based on the objective function in Eq. (\ref{eq:problem}a).}
	\end{algorithmic}
\end{algorithm}

\subsection{CRSC for Edge Video Transmission}
To enable efficient transmission of edge video while addressing the sensor’s resource constraints, we propose the CRSC scheme. Since the semantic encoder is deployed on the sensor, we utilize lightweight models such as BiFormer for semantic extraction and KAN for semantic encoding, thus minimizing the computational load on the encoder. Given the high level of redundancy in video frames, we design two SC models with distinct CRs to process static and dynamic frames separately. Additionally, to improve transmission quality, particularly under high compression, we employ a KD mechanism that facilitates mutual learning between the two SC models. The key modules of the CRSC scheme are detailed as follows:

\subsubsection{BiFormer-Based Semantic Extraction}
BiFormer, an innovative backbone based on ViT, adaptively focuses on relevant tokens while filtering out irrelevant ones, thereby achieving both high performance and computational efficiency \cite{zhu2023biformer}. We use BiFormer to extract semantic information from each video frame $\mathbf{x}_i$. As depicted in Fig. \ref{fig:SCCVS}, the key advantage of BiFormer over traditional ViT lies in its biLevel Routing Attention (BRA) mechanism, which operates as follows:

First, the input frame $\mathbf{x}_i$ is divided into $S \times S$ non-overlapping regions using a patch embedding layer, with each region containing $HW/S^2$ feature vectors. This process reshapes $\mathbf{x}_i$ into $\mathbf{x}^r_i \in \mathbb{R}^{S^2 \times HW/S^2 \times C}$. Linear projections are then applied to obtain query, key, and value tensors, denoted as $\mathbf{Q}$, $\mathbf{K}$, and $\mathbf{V} \in \mathbb{R}^{S^2 \times HW/S^2 \times C}$, respectively: \begin{equation}\label{eq:Bi1}
    \mathbf{Q}=\mathbf{x}^r_i \mathbf{W}^{q}, \quad \mathbf{K}=\mathbf{x}^r_i \mathbf{W}^{k}, \quad \mathbf{V}=\mathbf{x}^r_i \mathbf{W}^{v},
\end{equation}
where $\mathbf{W}^{q}$, $\mathbf{W}^{k}$, and $\mathbf{W}^{v}$ are the projection weights for the query, key, and value, respectively.

Second, region-level queries and keys, $\mathbf{Q}^r$ and $\mathbf{K}^r \in \mathbb{R}^{S^2 \times C}$, are computed by averaging the query and key tensors over each region. Using these, we calculate the adjacency matrix $\mathbf{A}^r \in \mathbb{R}^{S^2 \times S^2}$ to quantify semantic relationships between regions: 
\begin{equation}\label{eq:Bi2}
\mathbf{A}^{r}=\mathbf{Q}^{r}\left(\mathbf{K}^{r}\right)^\text{T}.
\end{equation}

The adjacency matrix is then pruned by retaining the top-$k$ semantic connections for each region, yielding the routing index matrix $\mathbf{I}^r \in \mathbb{N}^{S^2 \times k}$: 
\begin{equation}\label{eq:Bi3}
    \mathbf{I}^r = \operatorname{topk}(\mathbf{A}^r),
\end{equation}
where $\operatorname{topk}(\cdot)$ is the row-wise top-$k$ selection operator. Hence, the $i$-th row of $\mathbf{I}^r$ contains the indices of the k most semantically relevant regions for the $i$-th region.

Next, with the region-to-region routing index matrix $\mathbf{I}^r$, fine-grained token-to-token attention is performed by gathering the corresponding key and value tensors:
\begin{equation}\label{eq:Bi4}
    \mathbf{K}^{g}=\operatorname{gather}\left(\mathbf{K}, \mathbf{I}^{r}\right), \quad \mathbf{V}^{g}=\operatorname{gather}\left(\mathbf{V}, \mathbf{I}^{r}\right),
\end{equation}
where $\mathbf{K}^g$ and $\mathbf{V}^g$ are the gathered key and value tensors. Attention is then applied to the gathered key-value pairs, producing the output tensor:
\begin{equation}\label{eq:Bi5}
    \mathbf{O}=\operatorname{Attention}\left(\mathbf{Q}, \mathbf{K}^{g}, \mathbf{V}^{g}\right)+\operatorname{LCE}(\mathbf{V}),
\end{equation}
\begin{equation}\label{eq:Bi6}
    \operatorname{Attention}\left(\mathbf{Q}, \mathbf{K}^{g}, \mathbf{V}^{g}\right)=\operatorname{softmax}\left(\frac{\mathbf{Q}{\mathbf{K}^g}^\text{T}}{\sqrt{C}}\right) \mathbf{V}^g,
\end{equation}
where $\sqrt{C}$ is the scaling factor \cite{vaswani2017attention}, and $\operatorname{LCE}(\mathbf{V})$ refers to a local context enhancement term \cite{ren2022shunted}.

Finally, a linear projection layer $F_\text{Proj}$ is applied to the output tensor to obtain a high-dimensional semantic representation:
\begin{equation}\label{eq:Bi7}
    \mathbf{s}_i=F_\text{Proj}(\mathbf{O}).
\end{equation}

\subsubsection{KAN-Based Semantic Encoding}
In conventional SC models, the semantic encoding is typically performed using CNNs or FC layers, which map the semantic representation $\mathbf{s}_i$ to the semantic encoding $\mathbf{e}_i$. However, due to the dense nature of FC layers, these networks have a large number of parameters, leading to high computational costs for resource-constrained sensors. To address this, we employ the KAN, which efficiently approximates arbitrary multivariable functions through multi-level nonlinear compositions, requiring fewer parameters by utilizing a series of nested nonlinear functions \cite{liu2024kan}. The steps for semantic encoding using KAN are as follows:

First, the semantic representation $\mathbf{s}_i$ is flattened into an $n$-dimensional vector, $\mathbf{s}_i^\text{F}=[\mathbf{s}_{i,1}^\text{F},\mathbf{s}_{i,2}^\text{F},\dots, \mathbf{s}_{i,n}^\text{F}] \in \mathbb{R}^n$. Based on the Kolmogorov–Arnold theorem, any continuous function $f(\cdot)$ can be represented as: 
\begin{equation}\label{eq:KAN1}
    f(\mathbf{s}_{i,1}^\text{F},\mathbf{s}_{i,2}^\text{F},\dots, \mathbf{s}_{i,n}^\text{F}) = \sum_{q=1}^{2n+1} \varphi_q\left( \sum_{p=1}^n \psi_{qp}(\mathbf{s}_i^\text{F}) \right),
\end{equation}
where \(\varphi_q\) and \(\psi_{qp}\) are learnable nonlinear functions used for feature mapping and combination. Eq. (\ref{eq:KAN1}) provides the theoretical basis for KAN by demonstrating that any \(n\)-dimensional function can be constructed through a set of one-dimensional functions.

Next, KAN applies a nonlinear mapping to each component $\mathbf{s}_i^\text{F}$, generating intermediate feature representations. This mapping is performed by the function
\(\psi_{qp}\), defined as:
\begin{equation}\label{eq:KAN2}
    h_{qp} = \psi_{qp}(\mathbf{s}_i^\text{F}),
\end{equation}
where \(h_{qp}\) represents the mapped output for each input component $x_p$. This mapping is applied for each dimension
 \(p = 1, 2, \dots, n\) producing distinct intermediate features \(h_{qp}\). These intermediate features are then combined to form the final output representation. For each $q$, the combined feature $u_q$ is computed as:
\begin{equation}\label{eq:KAN3}
    u_q = \sum_{p=1}^n h_{qp}.
\end{equation}

Finally, a set of nonlinear functions, \(\varphi_q\) is applied to the combined features \(u_q\), producing the final semantic encoding:
\begin{equation}\label{eq:KAN4}
    \mathbf{e}_i = \sum_{q=1}^{2n+1} \varphi_q(u_q) = \sum_{q=1}^{2n+1} \varphi_q\left( \sum_{p=1}^n \psi_{qp}(\mathbf{s}_i^\text{F}) \right),
\end{equation}
where $\mathbf{e}_i \in \mathbb{R}^{m\cdot r_i}$ represents the semantic encoding based on the given CR $r_i$. After modulating on $\mathbf{e}_i$, the complex vector $\mathbf{c}_i$ is obtained and transmitted over the wireless channel, to be decoded at the receiver.

The inference phase of CRSC is summarized in \textbf{Algorithm \ref{alg:CRSC1}}.

\begin{algorithm}
\caption{Inference phase of CRSC}
\label{alg:CRSC1}
\begin{algorithmic}[1]
	\REQUIRE $\mathbf{x}_i$, $r_i$.
	\ENSURE $\mathbf{\hat{x}}_i$.
	\STATE{Extract the semantic representation $\mathbf{s}_i$ using Eqs. (\ref{eq:Bi1})-(\ref{eq:Bi7}).}
    \STATE{Perform the semantic encoding to obtain $\mathbf{e}_i$ using Eqs. (\ref{eq:KAN1})-(\ref{eq:KAN4}) with the given CR $r_i$.}
    \STATE{Generate the complex vector $\mathbf{c}_i$ by modulating $\mathbf{e}_i$.}
    \STATE{Receive the complex vector $\mathbf{y}_i$ according to Eq. (\ref{eq:SC3}).}
    \STATE{Reconstruct the frame $\mathbf{\hat{x}}_i$ using Eq. (\ref{eq:SC4}).} 
\end{algorithmic}
\end{algorithm}

\subsubsection{KD-Based Model Training}
To account for the varying nature of video content, we design two distinct SC models: one with a high CR, $r_\text{high}$, to process static frames, and another with a low CR, $r_\text{low}$, to process dynamic frames. Given that high CR can lead to significant semantic loss, we introduce KD during the training process to improve the performance of the high-CR SC model.
KD is a transfer learning approach technique that utilizes a mentor-student framework to transfer knowledge from a well-performing mentor model to a less capable student model. In this context, the low-CR SC model serves as the mentor, while the high-CR SC model acts as the student. The training process for the mentor and student models using KD is illustrated in Fig. \ref{fig:KD}, and is described as follows:
\begin{figure}[htbp]
	\centering
	\includegraphics[width=4.5cm]{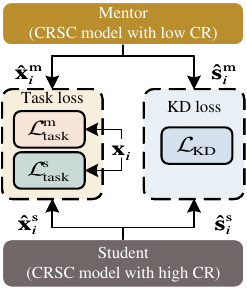}
	\caption{Illustration of the KD-based model training process.}
	\label{fig:KD}
\end{figure}

 \textbf{Distill Knowledge from Hard Labels:} Both the mentor and student models calculate the loss between their outputs and the corresponding hard labels, which are defined by the specific task at hand \cite{jafari2021annealing}. In this case, since the focus is on video frame reconstruction, the hard labels correspond to the original video frames. Let the input frame be $\mathbf{x}_i$; the task losses for the mentor and student models are defined as follows: 
\begin{equation}\label{eq:KD1}
\mathcal{L}_\text{task}^\text{m}=\operatorname{MSE}(\mathbf{x}_i,\mathbf{\hat{x}}_i^\text{m}), 
\end{equation}
\begin{equation}\label{eq:KD2}
    \mathcal{L}_\text{task}^\text{s}=\operatorname{MSE}(\mathbf{x}_i,\mathbf{\hat{x}}_i^\text{s}),
\end{equation}
where $\mathbf{\hat{x}}_i^\text{m}$ and $\mathbf{\hat{x}}_i^\text{s}$ denote the reconstructed frames generated by the mentor and student models after transmission, respectively. The mean-square error function, $\operatorname{MSE}(\cdot)$, is employed to ensure pixel-level consistency between the original frame and the reconstructed frames. In summary, these task losses provide direct task-specific supervision to guide both the mentor and student models during training.

\textbf{Distill Knowledge from Soft Labels:} In addition to hard label distillation, knowledge transfer between the mentor and student models also occurs through soft labels, such as the semantic encodings produced by each model \cite{jafari2021annealing}. Given that incorrect prediction from either the mentor or student model could negatively impact the other during KD, we implement an adaptive approach that adjusts the distillation loss based on the quality of the predicted hard labels (i.e., Eqs. (\ref{eq:KD1}) and (\ref{eq:KD2})). The adaptive distillation loss for the student model is expressed as:
\begin{equation}\label{eq:KD3}
\mathcal{L}_\text{KD}=\frac{{\rm{KL}}(\mathbf{\hat{s}}_i^\text{s},\mathbf{\hat{s}}_i^\text{m})}{\mathcal{L}_\text{task}^\text{m}},
\end{equation}
where ${\rm{KL}}(\cdot)$ represents the Kullback–Leibler divergence, and $\mathbf{\hat{s}}_i^\text{s}$ and $\mathbf{\hat{s}}_i^\text{m}$ are the semantic representations reconstructed by the student and mentor models, respectively. Specifically, $\mathbf{\hat{s}}_i^\text{s}$ is generated from the mentor’s semantic encoding $\mathbf{e}_i^\text{m} \in \mathbb{R}^{m \cdot r_\text{low}}$, while $\mathbf{\hat{s}}_i^\text{m}$ is derived from the student’s semantic encoding $\mathbf{e}_i^\text{s} \in \mathbb{R}^{m \cdot r_\text{high}}$.

Both the mentor and student models are trained by minimizing a combination of task and KD losses using the Stochastic Gradient Descent (SGD) optimizer \cite{zhang2018improved}.  Here, $G$ denotes the number of training epochs, and $\mathcal{D}$ represents the training dataset. The parameters $\alpha_\text{m}$ and $\beta_\text{m}$ refer to the semantic encoder and decoder in the mentor model, while $\alpha_\text{s}$ and $\beta_\text{s}$ represent the corresponding parameters in the student model. The training process for CRSC is outlined in \textbf{Algorithm \ref{alg:CRSC2}}.

\begin{algorithm}
\caption{Training Phase of CRSC}
\label{alg:CRSC2}
\begin{algorithmic}[1]
	\REQUIRE $\mathcal{D}$.
	\ENSURE $\alpha_\text{m}$, $\beta_\text{m}$, $\alpha_\text{s}$, $\beta_\text{s}$.
	\FOR{each epoch in $G$}
	\FOR{each batch sample in $\mathcal{D}$}
	\STATE{Compute task losses $\mathcal{L}_\text{task}^\text{m}$ and $\mathcal{L}_\text{task}^\text{s}$ using Eqs. (\ref{eq:KD1}) and (\ref{eq:KD2}).}
	\STATE{Compute KD loss $\mathcal{L}_\text{dis}$ using Eq. (\ref{eq:KD3}).}
	\STATE{Update $\alpha_\text{m}$ and $\beta_\text{m}$ by minimizing $\mathcal{L}_\text{task}^\text{m}$ using SGD optimizer.}
	\STATE{Update $\alpha_\text{s}$ and $\beta_\text{s}$ by minimizing $\mathcal{L}_\text{task}^\text{s} + \mathcal{L}_\text{KD}$ using the SGD optimizer.}
	\ENDFOR
	\ENDFOR
\end{algorithmic}
\end{algorithm}

\subsection{OSMS for Edge Video Sensing}
The CRSC model lacks inherent sensing capabilities and therefore cannot autonomously determine whether a given frame is static or dynamic, making it unable to adjust the CR in real time. Traditional radar-based sensing systems require specialized equipment on the sensor, leading to high sensing costs. To address this limitation, we propose the OSMS scheme, which intelligently detects changes in scenes and guides the transmission process of CRSC. Specifically, for each video frame, we employ an object detection model, Yolov10, to locate movable targets. Following this, we use a semantic segmentation model, FastSAM, to isolate key pixel sets within the frame. Finally, we compare contextual information across frames to detect scene changes, allowing CR adjustments to be made dynamically in the CRSC system. The OSMS process is outlined in the following steps:

\subsubsection{Yolov10-Based Object Location}
Yolov10, an end-to-end object detection model, is optimized for real-time applications, delivering high frame rates without compromising accuracy.  Its lightweight design enables deployment on devices with limited computational resources \cite{wang2024yolov10}. Therefore, we use Yolov10 to identify and locate movable targets in each frame $\mathbf{x}_i$. The detection process is as follows:

First, the current frame $\mathbf{x}_i$ is processed to obtain feature maps, denoted as $\mathbf{F}_i \in \mathbb{R}^{H' \times W' \times C'}$, where $H'$, $W'$, and $C'$ are the dimensions of the feature maps. This process can be expressed as:
\begin{equation}\label{eq:yo1}
    \mathbf{F}_i = \operatorname{Conv}(\mathbf{x}_i),
\end{equation}
where $\operatorname{Conv}(\cdot)$ represents the convolution operation.

Next, each anchor box is parameterized by its width $w_a$ and height $h_a$, customized to fit the objects detected in the feature maps $\mathbf{F}_i$. For the $j$-th anchor box, Yolov10 predicts a bounding box offset $\mathbf{t}_j = (t_j^x, t_j^y, t_j^w, t_j^h)$ and a confidence score $c_j$. The offset $\mathbf{t}_j$ adjusts the anchor box to better fit the detected object, which is formulated as follows: 
\begin{equation}\label{eq:yo2}
   x_j = \sigma(t_j^x) + x_j^g, 
\end{equation}
\begin{equation}\label{eq:yo3}
    y_j = \sigma(t_j^y) + y_j^g,
\end{equation}
\begin{equation}\label{eq:yo4}
    w_j = w_a \exp(t_j^w), 
\end{equation}
\begin{equation}\label{eq:yo5}
    h_j = h_a \exp(t_j^h),
\end{equation}
where $(x_j^g, y_j^g)$ is the grid cell's top-left corner, $\sigma(\cdot)$ is the sigmoid function, and $\exp(\cdot)$ ensures that the width and height are dynamically scaled. This flexibility in bounding box regression significantly improves detection precision. 

Finally, the set $\mathcal{B}_i$ of all the detected boxes in $\mathbf{x}_i$ is as the output:
\begin{equation}\label{eq:yo6}
    \mathcal{B}_i = \{(x_j, y_j, w_j, h_j, c_j)|j \in \{1,2,\dots,O_i\}\},
\end{equation}
where $(x_j, y_j, w_j, h_j)$ represents the coordinates of the $j$-th box, $c_j$ is the confidence score for the detected object, and $O_i$ represents the number of detected objects. 
Thus, $\mathcal{B}_i$ provides the location of all the movable targets in the current frame $\mathbf{x}_i$.

\subsubsection{FastSAM-Based Semantic Segmentation}
The Segment Anything Model (SAM) \cite{kirillov2023segment}, built on a ViT architecture, incurs high computational overhead when processing high-resolution images, limiting its practical applicability in industrial settings. FastSAM, a lightweight version of SAM, addresses this issue by utilizing a CNN-based detector for segmentation tasks, significantly reducing computational costs while maintaining competitive performance. Therefore, we employ FastSAM to isolate the pixel sets of the targets detected by Yolov10.

In this approach, the frame \(\mathbf{x}_i\), along with the location data \(\mathcal{B}_i\) (obtained from Yolov10), is provided as input to FastSAM, denoted as \(F_{\Gamma}\), to perform segmentation \cite{jiang2024large}:
\begin{equation}\label{eq:fs1}
    F_{\Gamma}: (\mathbf{x}_i, \mathcal{B}_i) \rightarrow (\mathbf{M}_i, \mathbf{S}_i, \mathbf{L}_i),
\end{equation}
where $\mathbf{M}_i$ represents the generated binary mask with dimensions \((H, W)\), indicating whether a pixel belongs to the target object (1) or not (0). Additionally, $\mathbf{S}_i$ represents the Intersection over Union (IoU) score, measuring the overlap between the mask and the ground truth annotation, while $\mathbf{L}_i$ provides the class label of the detected object. The resulting mask $\mathbf{M}_i$ is treated as the sensing result for frame $\mathbf{x}_i$.

To detect changes between consecutive frames, we define the difference between the sensing results of two consecutive frames, \(\mathbf{M}_{i-1}\) and \(\mathbf{M}_i\), as follows:
\begin{equation}\label{eq:fs2}
     \eta_i = \frac{1}{HW}\sum_{a=1}^H \sum_{b=1}^W |\mathbf{M}_{i,a,b} - \mathbf{M}_{i-1,a,b}|,
\end{equation}
where \(\mathbf{M}_{i,a,b}\) denotes each pixel in the mask. Since \(\mathbf{M}_{i,a,b} \in \{0,1\}\), \( \eta_i < \epsilon \) indicates that there are no changes between frames, classifying \(\mathbf{M}_i\) as a static. $\epsilon$ represents a threshold, default as $1e-4$ in this paper. 
If changes are detected, the frame is labeled as dynamic. Based on this classification, the appropriate CR \(r_i \in \{r_\text{low}, r_\text{high}\}\) is assigned to the frame.
The workflow for OSMS is outlined in \textbf{Algorithm \ref{alg:OSMS}}.
\begin{algorithm}
\caption{OSMS}
\label{alg:OSMS}
\begin{algorithmic}[1]
	\REQUIRE $\mathcal{B}_i$, $\mathbf{x}_i$.
	\ENSURE $r_i$.
	\STATE{Obtain the location information $\mathcal{B}_i$ for frame $\mathbf{x}_i$ using Eqs. (\ref{eq:yo1})-(\ref{eq:yo6}).}
    \STATE{Based on $\mathcal{B}_i$, obtain the sensing results $\mathbf{M}_i$ using Eq. (\ref{eq:fs1}).}
    \STATE{Calculate the difference $\eta_i$ between the current and previous frames using Eq. (\ref{eq:fs2}).}
    \IF{$\eta_i==0$}
    \STATE{$r_i=r_\text{low}$.}
    \ELSE
    \STATE{$r_i=r_\text{high}$.}
    \ENDIF
\end{algorithmic}
\end{algorithm}

\section{Experimental Simulations}
This section describes the simulation dataset, parameter settings, and evaluation results. The simulations are conducted on a server equipped with an Intel Xeon CPU (2.4 GHz, 128 GB RAM) and an NVIDIA A800 GPU (80 GB SGRAM), utilizing the PyTorch framework to implement the proposed SC schemes. The results demonstrate the effectiveness of the proposed SCCVS framework in transmitting videos.

\subsection{Simulation Settings}
\subsubsection{Dataset Setup}
To evaluate the proposed methods, we employ the VIRAT Video Dataset \cite{oh2011large}, which contains a variety of surveillance videos captured in different scenarios. The dataset is divided into two primary activity categories: single-object and two-object scenarios, involving both humans and vehicles.

During the training phase, we capture one frame per second from each video, resulting in approximately 9,600 RGB images. This dataset is used to train the CRSC model as described in \textbf{Algorithm \ref{alg:CRSC2}}. 
For the inference phase, we select a continuous video segment as test data, which is processed according to \textbf{Algorithm \ref{alg:SCCVS}}. Notably, Given the limited number of static scenes in the dataset, we augment the test video by duplicating frames from static segments, ensuring that the ratio of static to dynamic frames in the test video is approximately 6:4.

\subsubsection{Parameters Settings}
For the system model, the bandwidth is set to $B = 1$ KHz, and the SNR is varied between 0 dB and 25 dB. In the inference phase, we assess the SC model using different fixed SNR values. During training, the SNR is randomly varied in each forward propagation to improve the robustness of the CRSC model against channel noise. In the inference phase, we evaluate the SC model under fixed SNR conditions.
Additionally, for the CR, we make the following simplifications: when $r_i = r_\text{low}$, the length of the semantic encoding $\mathbf{e}_i$ is set to 256. Conversely, when $r_i = r_\text{high}$, the length of $\mathbf{e}_i$ is set to 16. This implies that for static frames, only $6.25\%$ of the data volume required for dynamic frames is transmitted.

For the CRSC model, we first employ the BiFormer to perform semantic extraction from the original frames, followed by the KAN for semantic encoding. Finally, a ViT-based decoder \cite{he2022masked} is employed for frame reconstruction. Table \ref{tab:CRSC} provides the detailed parameter settings for each module, highlighting the computational requirements at the receiver (i.e., the base station).
\renewcommand{\arraystretch}{1.5} 
\begin{table}[htbp]
	\centering
	\caption{Parameters of the CRSC model}
	\label{tab:CRSC}
	\begin{tabular}{|c|cc|c|}
\hline
\multirow{2}{*}{} Models & \multicolumn{2}{c|}{Semantic encoder}       & Semantic encoder \\ \cline{2-4} 
                  & \multicolumn{1}{c|}{BiFormer}   & KAN       & ViT decoder      \\ \hline
Parameters        & \multicolumn{1}{c|}{13,142,760} & 1,179,994 & 111,754,752      \\ \hline
\end{tabular}
\end{table}

For the OSMS scheme, we aim to achieve a lightweight model by utilizing Yolov10-n and FastSAM-s for target detection and semantic segmentation, respectively. Additionally, we consider an OSMS (w/o lightweight) scheme, which employs Yolov10-x and FastSAM-x. Table \ref{tab:OSMS} presents a description of their parameters.
\renewcommand{\arraystretch}{1.5} 
\begin{table}[htbp]
	\centering
	\caption{Parameters of the CRSC model}
	\label{tab:OSMS}
\begin{tabular}{|c|cc|cc|}
\hline
\multirow{2}{*}{} Models & \multicolumn{2}{c|}{OSMS}                   & \multicolumn{2}{c|}{OSMS (w/o lightweight)}  \\ \cline{2-5} 
                  & \multicolumn{1}{c|}{Yolov10-n} & FastSAM-s  & \multicolumn{1}{c|}{Yolov10-x}  & FastSAM-x  \\ \hline
Parameters        & \multicolumn{1}{c|}{2,775,520} & 11,790,483 & \multicolumn{1}{c|}{31,808,960} & 72,234,149 \\ \hline
\end{tabular}
\end{table}

\subsubsection{Comparison Schemes}
To demonstrate the superiority of the proposed SCCVS framework, we consider the following comparison schemes: 
\begin{itemize}
    \item SCCVS (w/o OSMS):  In this variant, the CRSC model is used for all frames without receiving CR prompts from the OSMS. As a result, all frames are transmitted with a low CR. 
    \item SCCVS (w/o KD): In this model, KD is excluded from the training phase, and the high-CR and low-CR models are trained independently.
    \item DeepJSCC-V \cite{zhang2023predictive}: A baseline image semantic communication model featuring predictive and adaptive deep coding. 
\end{itemize}

\subsubsection{Evaluation Metrics}
To evaluate the CV sensing performance, we use the IoU loss to quantify the difference between OSMS and OSMS (w/o lightweight). IoU is commonly used in object detection and segmentation tasks to measure the overlap between two regions. The IoU loss function is defined as:
\begin{equation}
    \mathcal{L}_\text{IoU}=1-\frac{O(\mathcal{B}_i, \mathcal{\hat{B}}_i)}{S(\mathcal{B}_i)+S(\mathcal{\hat{B}}_i)-O(\mathcal{B}_i, \mathcal{\hat{B}}_i)},
\end{equation}
where $\mathcal{B}_i$ and $\mathcal{\hat{B}}_i$ represent the bounding boxes detected by OSMS and OSMS (w/o lightweight) in the $i$-th frame. $O(\mathcal{B}_i, \mathcal{\hat{B}}_i)$ represents the overlap area between the two boxes, and $S(\mathcal{B}_i)$ and $S(\mathcal{\hat{B}}_i)$ are their respective areas. A lower value of $\mathcal{L}_\text{IoU}$ indicates closer agreement between the detection results of the two models.

To assess the performance of the SCCVS framework for transmitting videos, we use peak signal-to-noise ratio (PSNR), multi-scale structural similarity index (MS-SSIM), and deep learning-based perceptual loss \cite{zhang2018unreasonable} as metrics.
PSNR measures the quality of a reconstructed image and is typically expressed in decibels (dB), with higher values indicating better quality:
\begin{equation}
	\mathrm{PSNR}(\mathbf{x}_i,\mathbf{\hat{x}}_i) = 10 \cdot \log_{10} \left( \frac{\mathrm{MAX}_I^2}{\mathrm{MSE}(\mathbf{x}_i,\mathbf{\hat{x}}_i)} \right),
\end{equation}
where $\mathrm{MAX}_I$ denotes the maximum possible pixel value of the image, typically 255 for 8-bit images.
MS-SSIM is an extension of the structural similarity index (SSIM) that evaluates the perceived quality of images by considering changes across multiple scales. It can be calculated as follows:
\begin{equation}
    \text{MS-SSIM}(\mathbf{x}_i, \mathbf{\hat{x}}_i) = \prod_{j=1}^{J} \text{SSIM}(\mathbf{x}_i, \mathbf{\hat{x}}_i)^{\alpha_j},
\end{equation}
\begin{equation}
	\mathrm{SSIM}(\mathbf{x}_i,\mathbf{\hat{x}}_i) = \frac{(2\varphi_{\mathbf{x}_i}\varphi_{\mathbf{\hat{x}}_i} + c_1)(2\phi_{\mathbf{x}_i\mathbf{\hat{x}}_i} + c_2)}{(\varphi_{\mathbf{x}_i}^2 + \varphi_{\mathbf{\hat{x}}_i}^2 + c_1)(\phi_{\mathbf{x}_i}^2 + \phi_{\mathbf{\hat{x}}_i}^2 + c_2)},
\end{equation}
where $J$ is the number of scales, $\alpha_j$ are weights assigned to each scale, typically summing to 1, $\varphi_{\mathbf{x}_i}$ and $\varphi_{\mathbf{\hat{x}}_i}$ are their means, $\phi_{\mathbf{x}_i}^2$ and $\phi_{\mathbf{\hat{x}}_i}^2$ are their variances, $\phi_{\mathbf{x}_i\mathbf{\hat{x}}_i}$ is their covariance, $c_1$ and $c_2$ are two constants used to avoid division by zero. Finally, the perceptual loss is calculated by using deep learning models (e.g., VGG-16 or ResNet-50) to obtain latent representations of $\mathbf{x}_i$ and $\mathbf{\hat{x}}_i$. The perceptual loss is the mean squared error between these representations, which can be expressed as follows:
\begin{equation}
    \mathcal{L}_\text{perception}=\operatorname{MSE}(P(\mathbf{x}_i),P(\mathbf{\hat{x}}_i)),
\end{equation}
where $P(\cdot)$ represents the perception network (e.g., VGG-16 or ResNet-50) used for feature extraction.

\subsection{Evaluation of CV Sensing}
This subsection presents the simulations for CV sensing, aiming to show the effectiveness of CV sensing and demonstrate the advantages of the OSMS over the OSMS (w/o lightweight).

\begin{figure}[htbp]
	\centering
	\includegraphics[width=8.5cm]{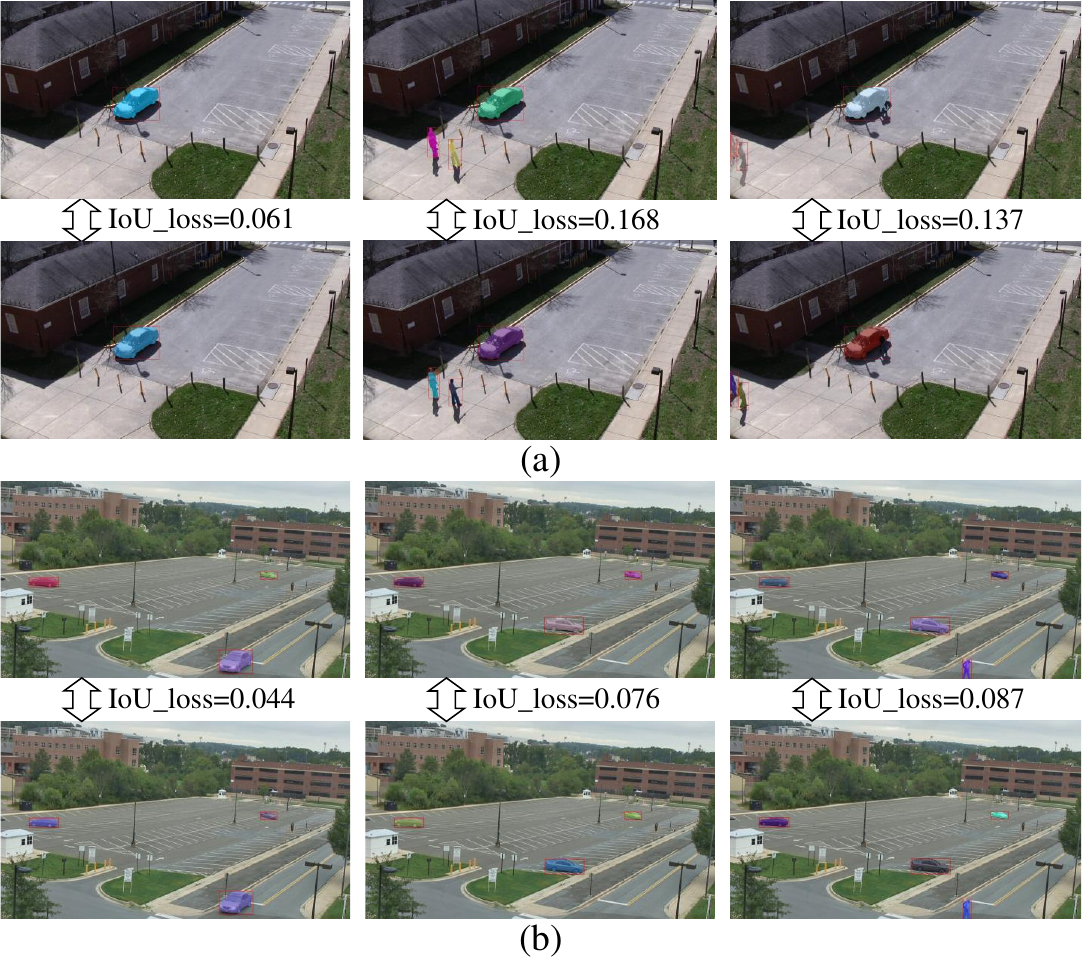}
	\caption{Comparison of the sensing results between the OSMS and the OSMS (w/o lightweight) in two scenes (a) and (b). The top result corresponds to OSMS, while the bottom corresponds to OSMS (w/o lightweight).}
	\label{exp:sensing_1}
\end{figure}

\begin{figure}[htbp]
	\centering
	\includegraphics[width=8cm]{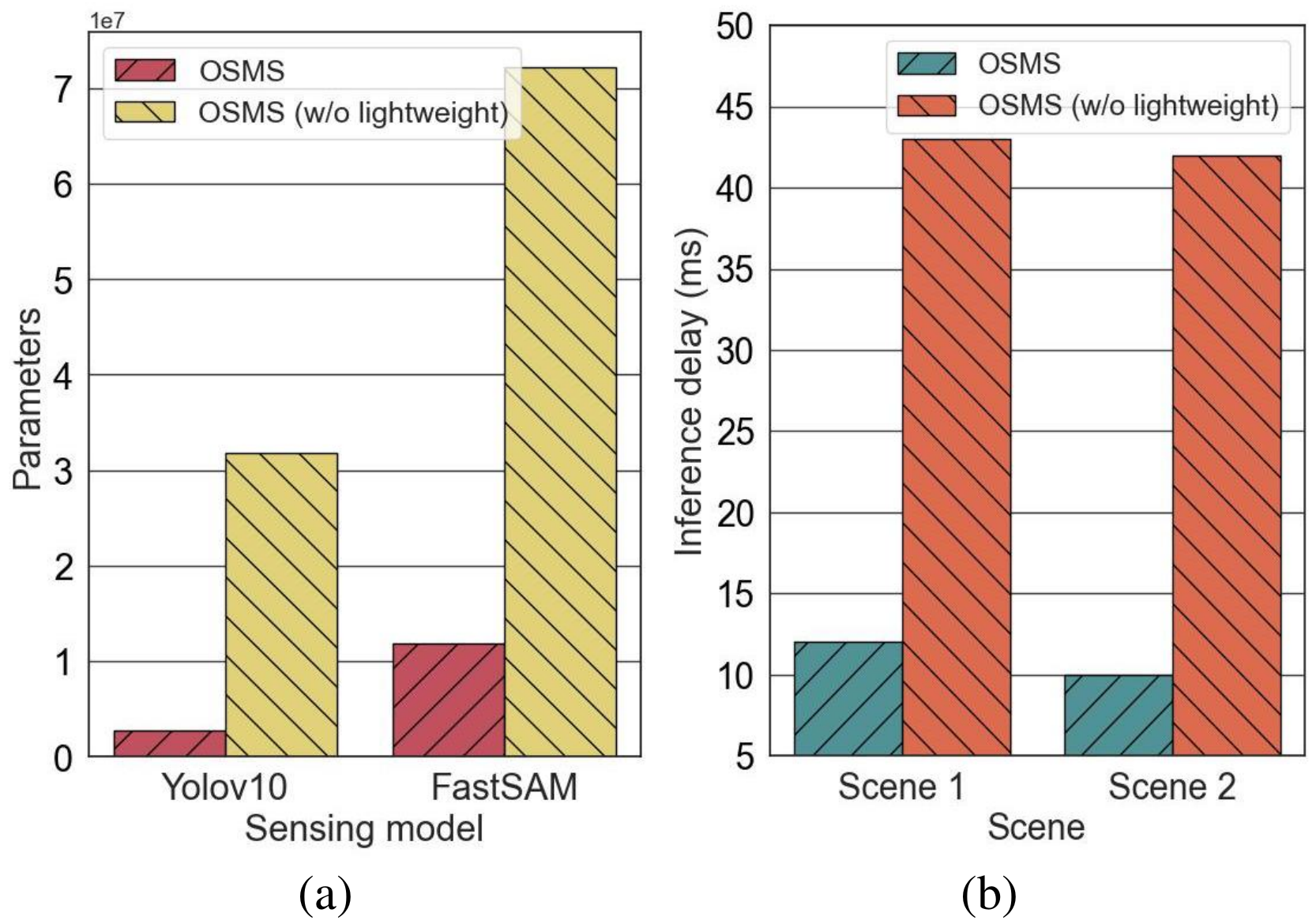}
	\caption{Evaluation of lightweight model performance in OSMS. (a) Parameter comparison between Yolov10 and FastSAM models in OSMS and OSMS (w/o lightweight). (b) Inference time comparison between OSMS and non-lightweight OSMS in two scenes.}
	\label{exp:sensing_2}
\end{figure}

As shown in Fig. \ref{exp:sensing_1}, we illustrate the sensing outcomes for different frames in two scenarios, emphasizing the performance differences between OSMS and OSMS (w/o lightweight). Both schemes accurately detect key objects, such as cars and pedestrians. To further quantify the differences, we employ Intersection over Union (IoU) loss, which shows that the sensing results from both methods are quite similar, particularly when the number of detected key objects is low.
Fig. \ref{exp:sensing_2} compares the parameters and inference times between OSMS and OSMS (w/o lightweight). The results indicate that OSMS has approximately $70\%$ fewer parameters and achieves inference speeds nearly four times faster than the non-lightweight version.

These findings demonstrate that the proposed OSMS achieves real-time, accurate sensing while maintaining a significantly lower computational load. This can be attributed to the robustness of the employed CV model, ensuring reliable performance. Additionally, the predominantly static nature of the edge scenes used in this study simplifies the sensing task, enabling even a lightweight CV model to produce accurate results.

\subsection{Evaluation of Video Transmission}
This subsection presents the simulations for video transmission via SC, focusing on the advantages of the proposed SCCVS framework in terms of transmission accuracy and delay.

\begin{figure}[htbp]
	\centering
	\includegraphics[width=8cm]{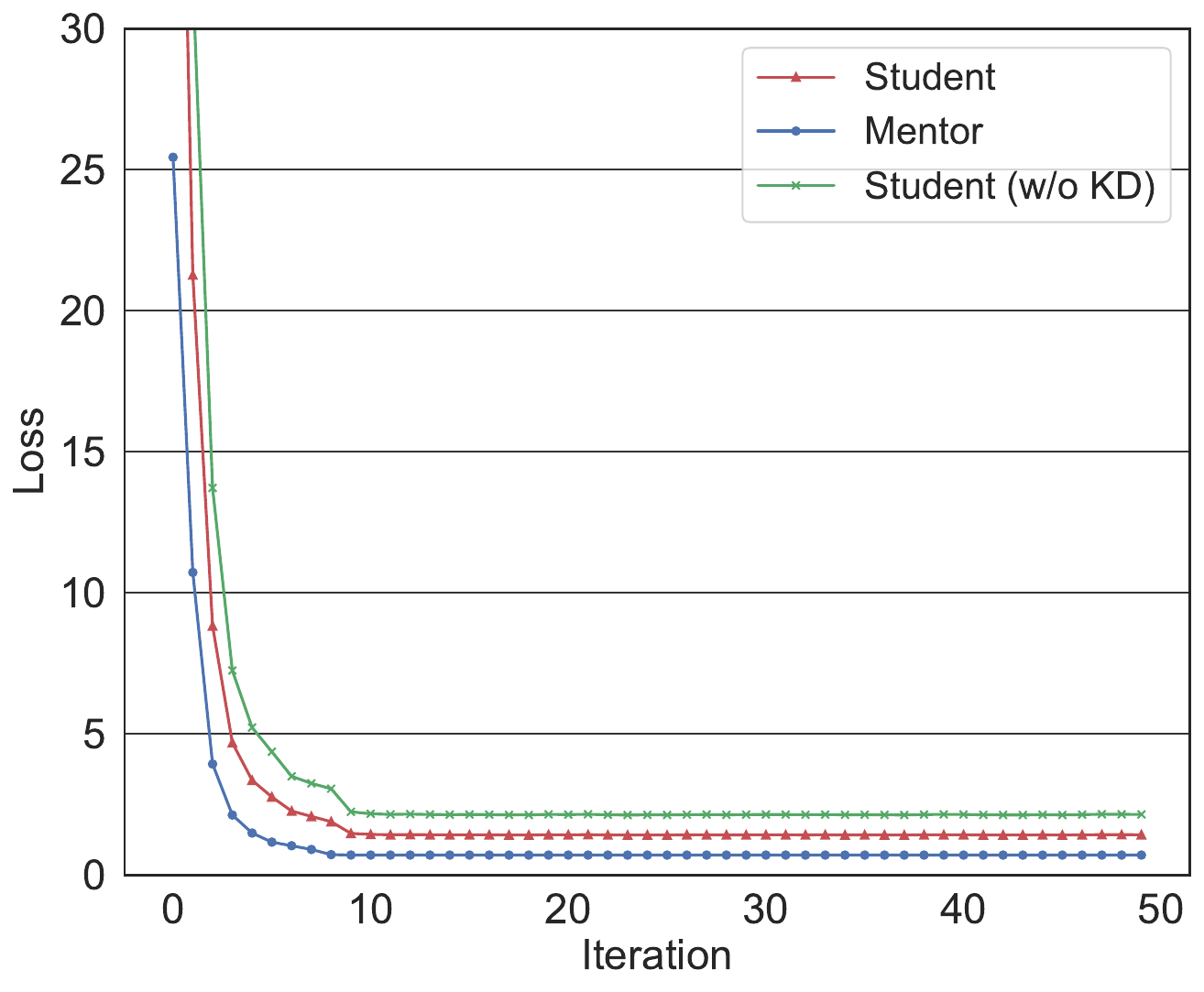}
	\caption{Training results of different schemes.}
	\label{exp:sc_loss}
\end{figure}

Fig. \ref{exp:sc_loss} compares the training losses of various models. The CRSC model with a low CR (acting as the mentor) achieves the fastest convergence, followed by the CRSC model with a high CR (the student). The student model without KD performs the worst. This difference is attributed to the mentor model transmitting richer semantic information, enabling its decoder to perform more accurate semantic decoding. The student model, though performing moderately due to transmitting fewer semantic features, still benefits from KD. In contrast, the student model without KD fails to learn from the mentor, leading to suboptimal performance. These results highlight the effectiveness of KD-based training.

\begin{figure}[htbp]
	\centering
	\includegraphics[width=8.5cm]{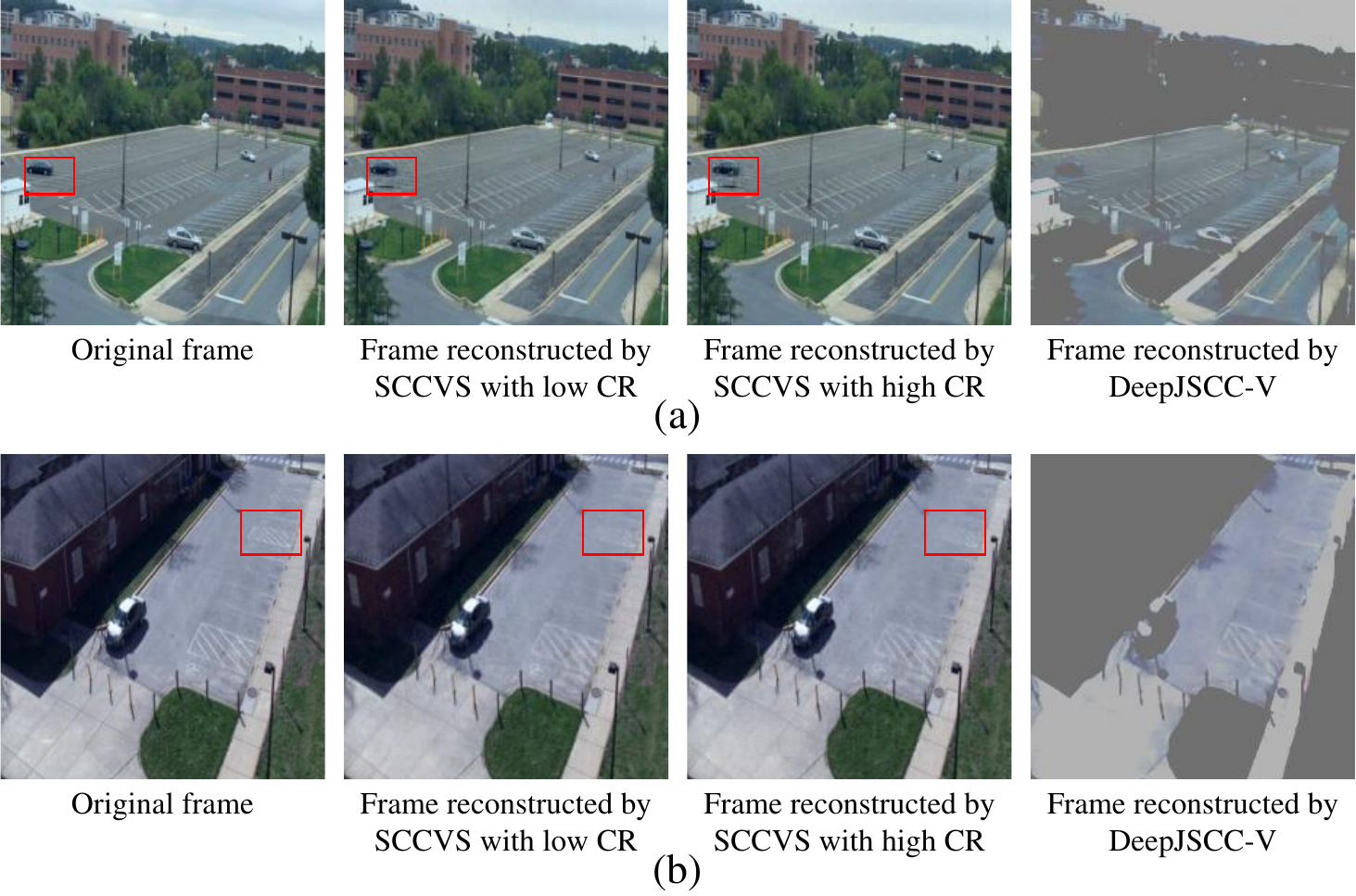}
	\caption{Comparison of the SC results with different schemes on two scenes (a) and (b) under SNR is 5 dB.}
	\label{exp:sc_1}
\end{figure}

Fig. \ref{exp:sc_1} visually compares the reconstructed frames under different SC schemes at an SNR of 5 dB. The frames generated by SCCVS closely resemble the original, with only minor detail loss, while those reconstructed by DeepJSCC-V suffer from significant color distortions and missing details. This highlights the advantage of SCCVS, which utilizes a lightweight encoder and deep decoder architecture within the CRSC model, optimizing reconstruction quality while reducing transmitter overhead.

\begin{figure}[htbp]
	\centering
	\includegraphics[width=8cm]{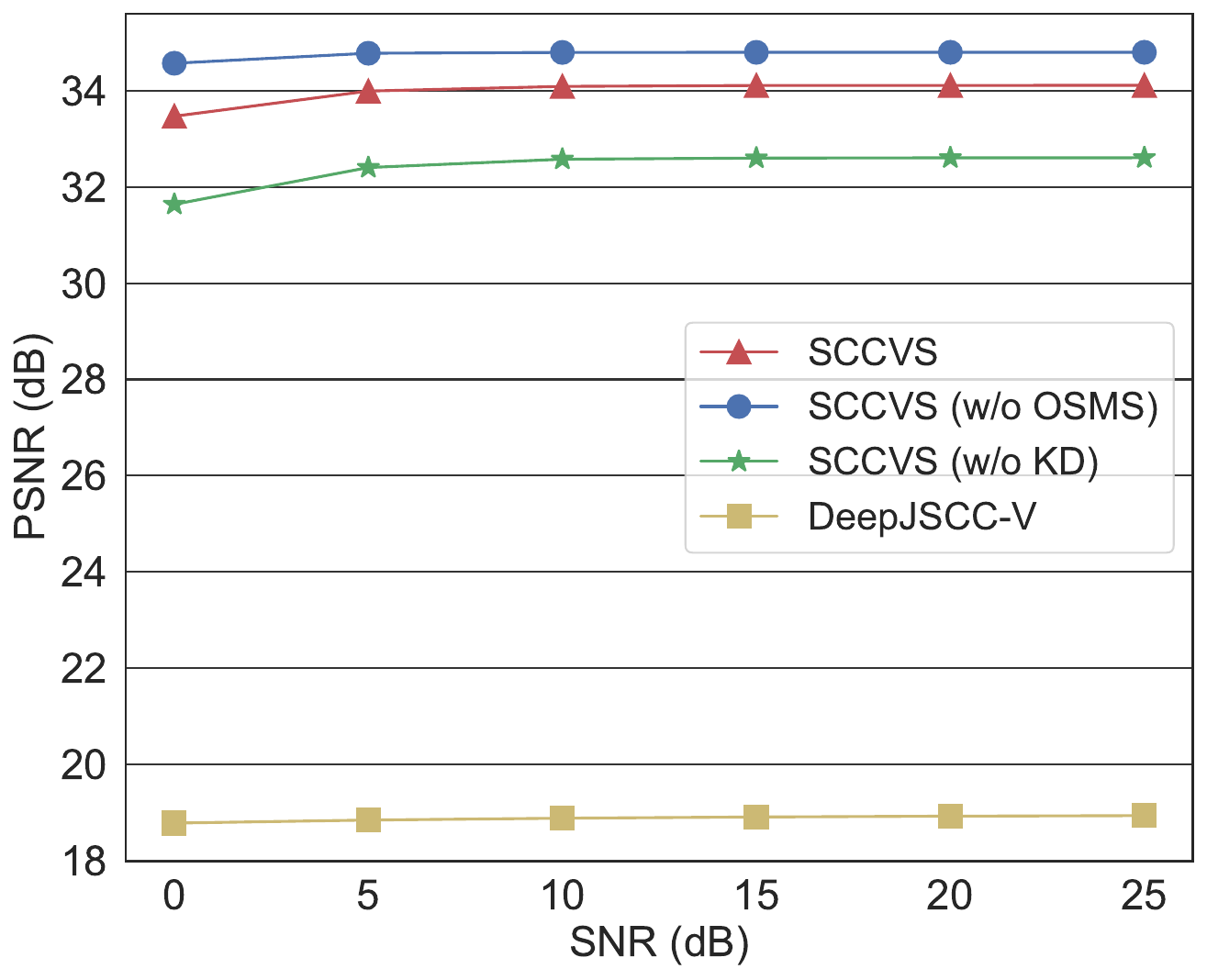}
	\caption{PSNR comparisons across different schemes.}
	\label{exp:sc_psnr}
\end{figure}

\begin{figure}[htbp]
	\centering
	\includegraphics[width=8cm]{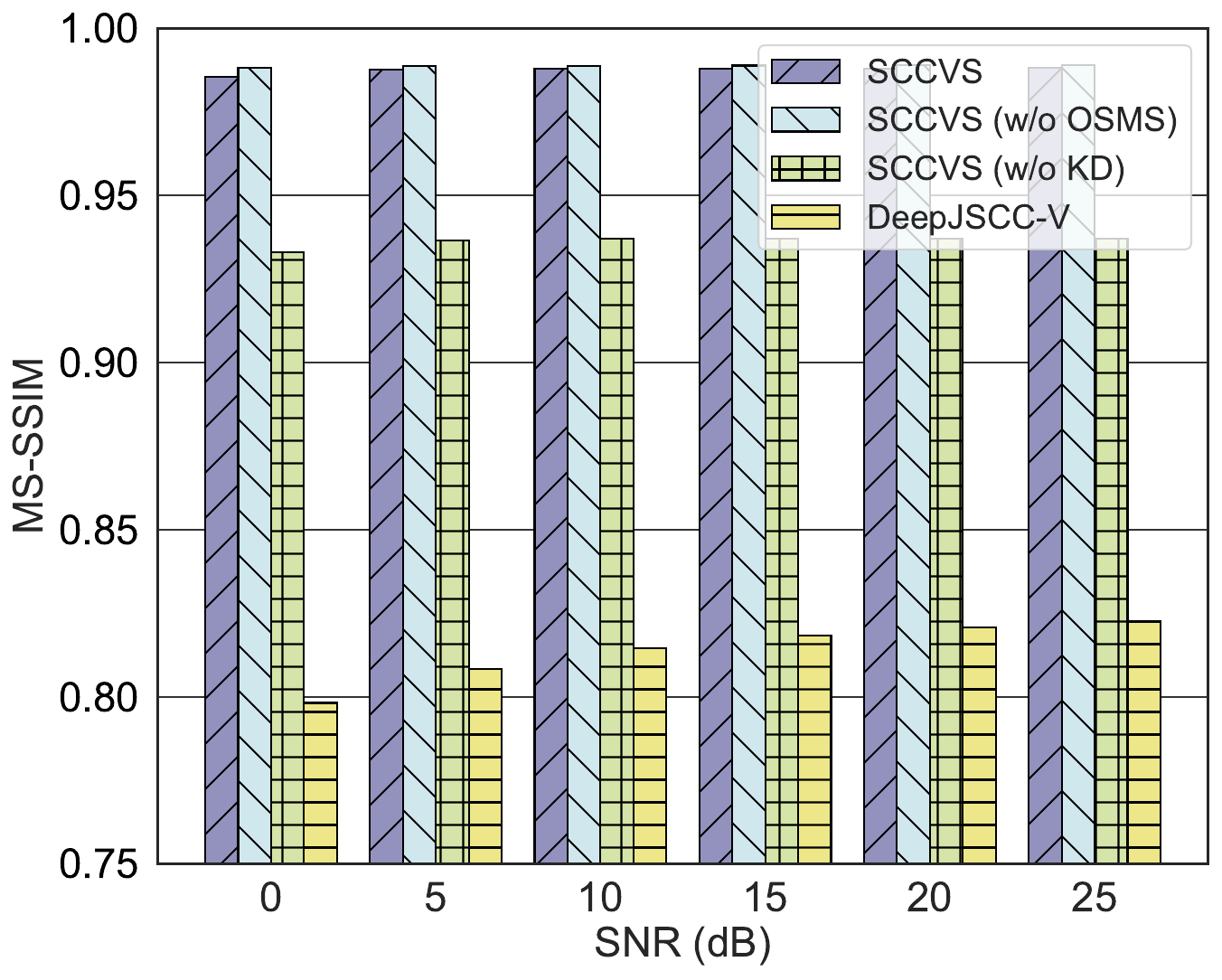}
	\caption{MS-SSIM comparisons across different schemes.}
	\label{exp:sc_ssim}
\end{figure}

\begin{figure}[htbp]
	\centering
	\includegraphics[width=8cm]{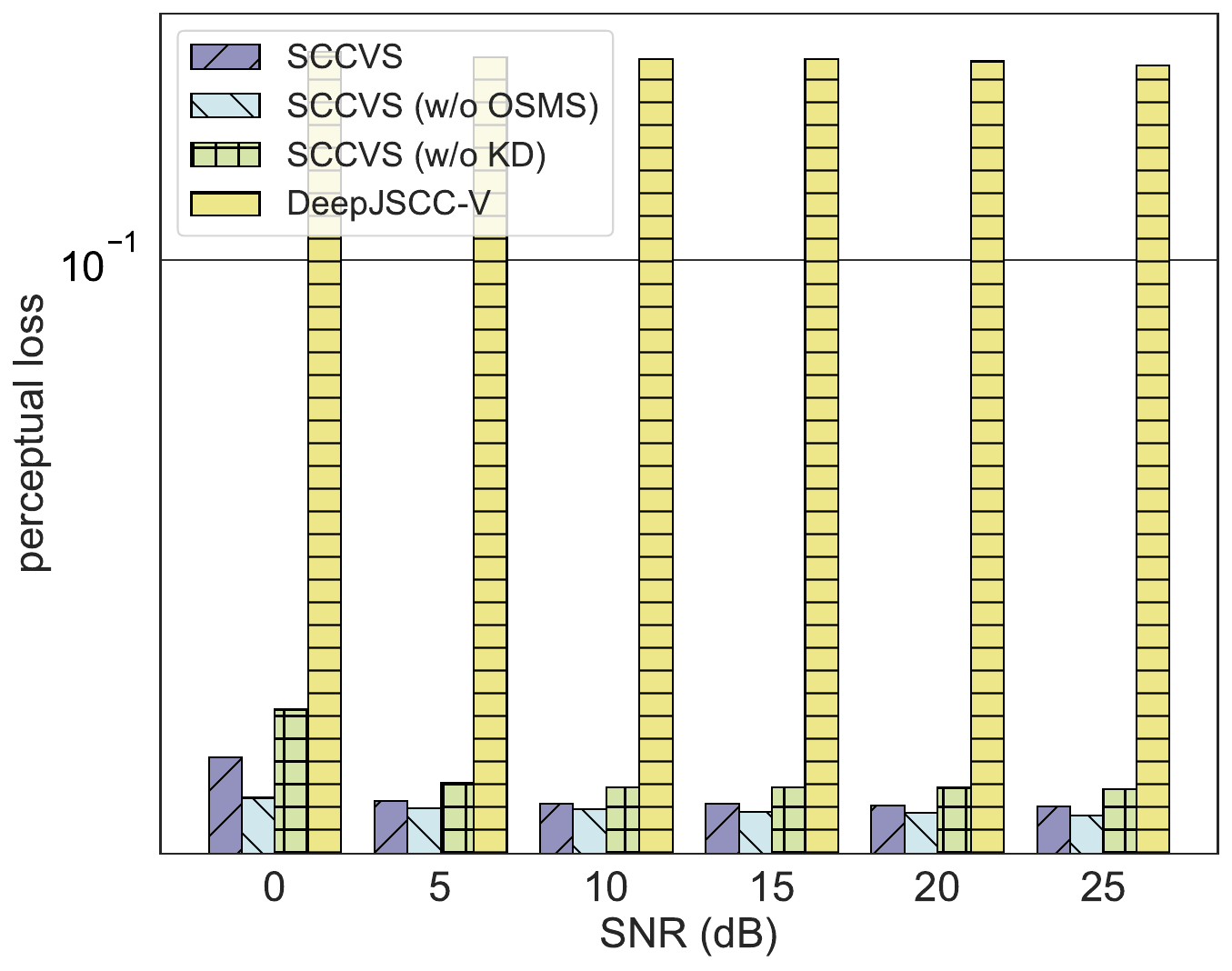}
	\caption{Perceptual loss comparisons across different schemes based on VGG-16.}
	\label{exp:sc_ploss}
\end{figure}

\begin{figure}[htbp]
	\centering
	\includegraphics[width=8cm]{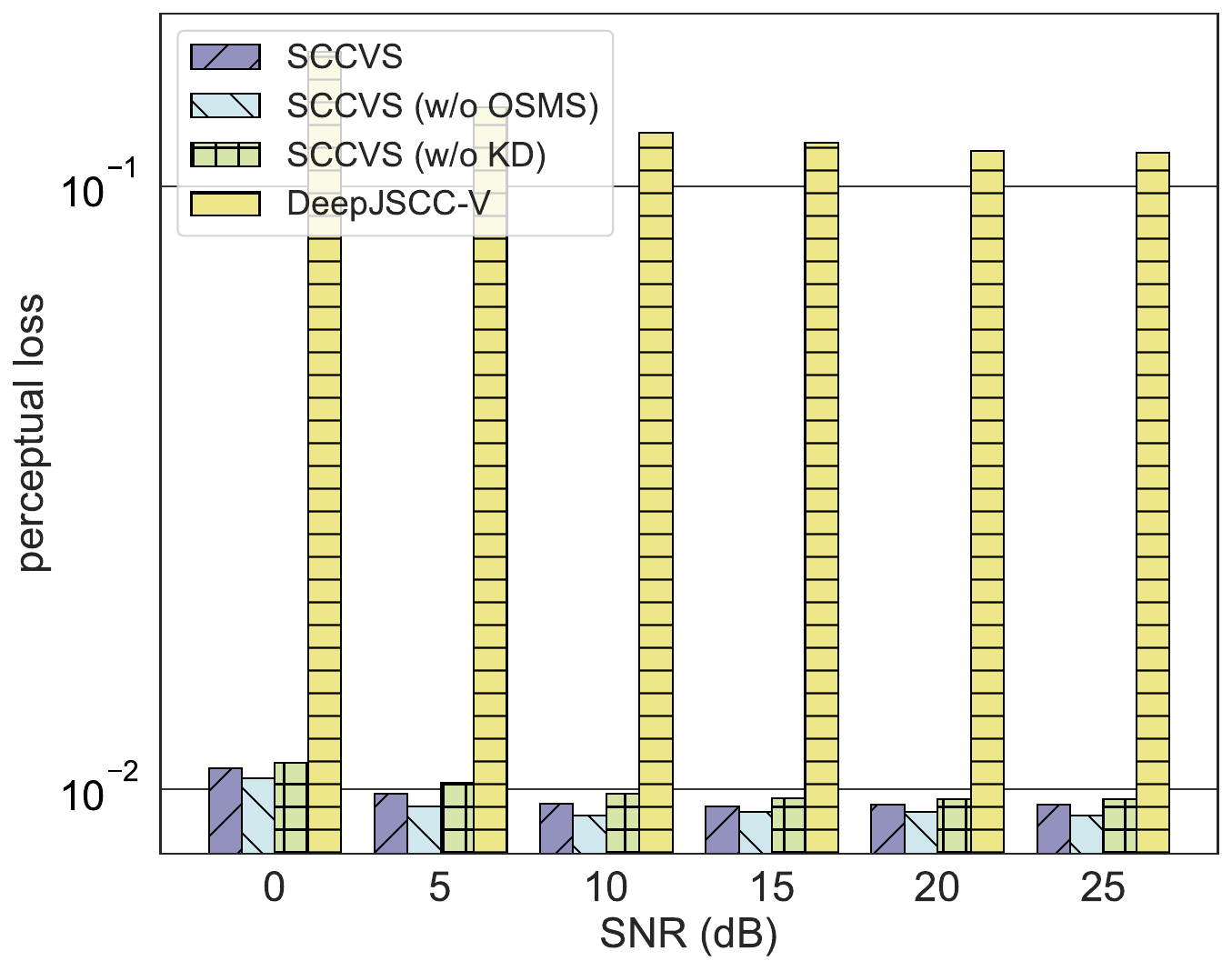}
	\caption{Perceptual loss comparisons across different schemes based on ResNet-50.}
	\label{exp:sc_ploss2}
\end{figure}

Figs. \ref{exp:sc_psnr} and \ref{exp:sc_ssim} compare the PSNR and MS-SSIM results across various schemes. SCCVS and SCCVS (w/o OSMS) achieve superior performance, while SCCVS (w/o KD) and DeepJSCC-V perform worse. Notably, SCCVS (w/o OSMS), which transmits all frames with a low CR, achieves higher accuracy. However, SCCVS strikes a balance between compression and accuracy, while the absence of KD in SCCVS (w/o KD) limits its performance, especially when transmitting static frames.

Figs. \ref{exp:sc_ploss} and \ref{exp:sc_ploss2} display the perceptual loss results based on VGG-16 and ResNet-50, respectively, with similar trends observed across both networks. Notably, using ResNet-50 reveals more pronounced differences between the schemes, further illustrating that SCCVS effectively minimizes visual distortion. 

\begin{figure}[htbp]
	\centering
	\includegraphics[width=8cm]{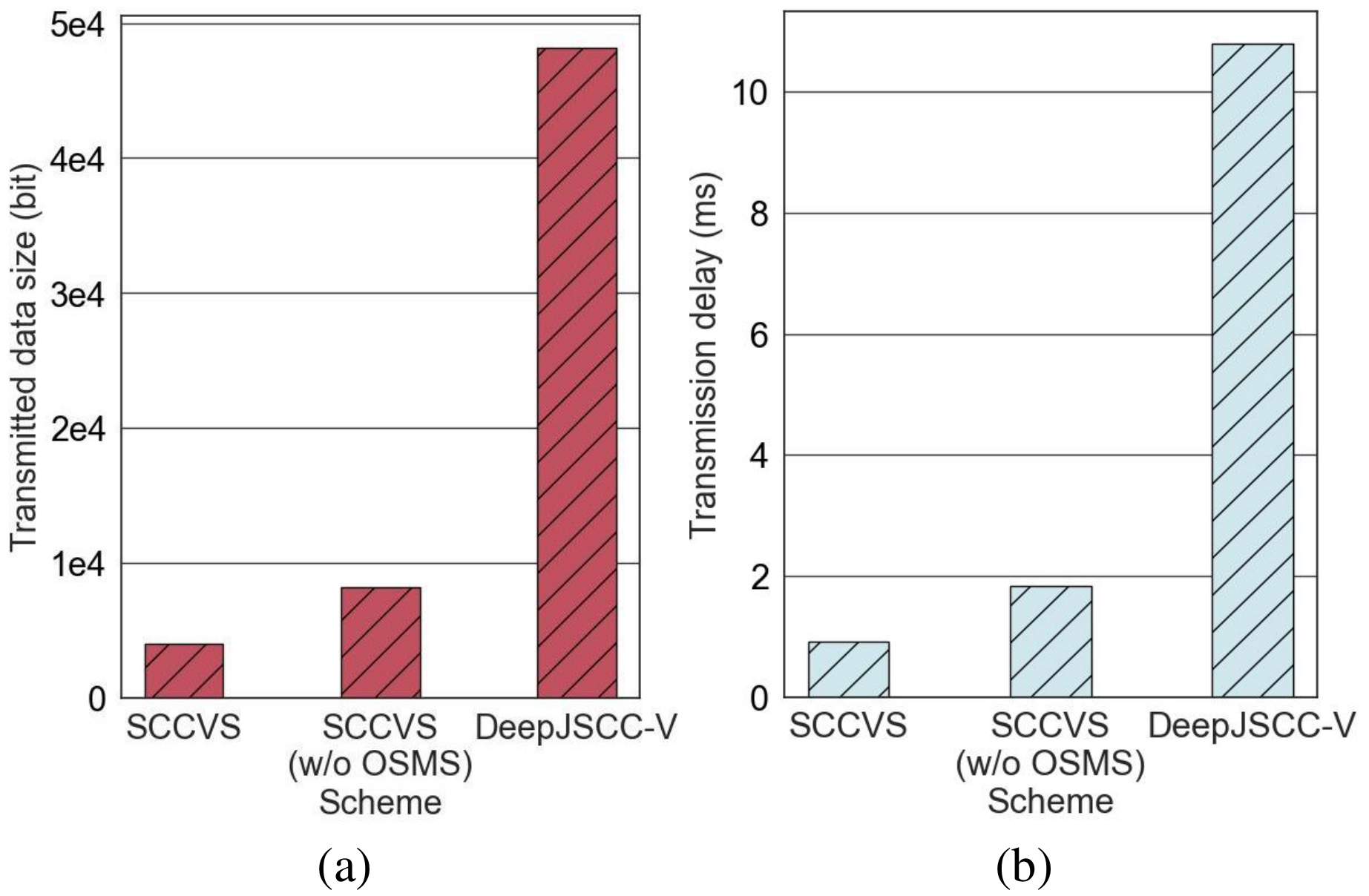}
	\caption{Comparison of the three schemes in terms of data transmission. (a) Comparison of the transmitted data size. (b) Comparison of the transmission delay.}
	\label{exp:sc_2}
\end{figure}

Fig. \ref{exp:sc_2} compares the transmission data size and delay. SCCVS significantly outperforms DeepJSCC-V in both data compression and transmission rate. However, SCCVS (w/o OSMS), which transmits all frames without adaptive sensing, incurs higher data transmission and longer delays due to the lack of CR adjustment.

\section{Conclusions}
To mitigate the high spectrum resource demands associated with vision sensors transmitting edge video, we propose the SCCVS framework. This framework introduces a CRSC model that intelligently adjusts the CRs of video frames based on real-time sensing results, optimizing the balance between compression efficiency and semantic fidelity. Additionally, the OSMS scheme is incorporated, leveraging CV techniques to detect changes in the edge scenes and assess the contextual importance of each frame. This enables OSMS to guide the CRSC model in applying lower CRs to dynamic frames while assigning higher CRs to static frames.

Furthermore, both the CRSC and OSMS models are designed with lightweight architectures, making the framework particularly suitable for resource-constrained sensors in real-world edge applications. Experimental simulations demonstrate a substantial reduction in bandwidth consumption, confirming the SCCVS framework's effectiveness in transmitting edge video without compromising critical semantic information.

Future work will focus on deploying the proposed system and algorithms on real hardware platforms, such as NVIDIA Jetson, to further validate the framework's practicality and performance in real-world scenarios.

\bibliography{bare_jrnl}
\bibliographystyle{IEEEtran}

\newpage
\end{document}